\documentclass[10pt,twocolumn,letterpaper]{article}
\pdfoutput=1
\usepackage{iccv}              
\usepackage{amsthm}
\usepackage{multirow}
\usepackage[table]{xcolor}
\usepackage{booktabs}
\usepackage{csquotes}
\definecolor{iccvblue}{rgb}{0.21,0.49,0.74}
\usepackage[pagebackref,breaklinks,colorlinks,allcolors=iccvblue]{hyperref}
\usepackage[capitalize]{cleveref}

\newtheorem{definition}{Definition}

\crefname{thm}{theorem}{theorems}
\crefname{defn}{definition}{definitions}
\crefname{lem}{lemma}{lemmas} 
\crefname{cor}{corollary}{corollaries} 
\crefname{conj}{conjecture}{conjectures}
\crefname{prop}{proposition}{propositions}
\crefname{alg}{algoirithm}{algorithms}
\def\paperID{5031} 
\def\confName{ICCV}
\def\confYear{2025}

\newcommand{\name}[0]{NCPTM-CIL\xspace}

\renewcommand{\vec}[1]{\boldsymbol{#1}}

\title{Enhancing Pre-Trained Model-Based Class-Incremental Learning through Neural Collapse}

\author{Kun He\thanks{Equal contribution to this work.}\textsuperscript{\rm ~~1}~~~Zijian Song\footnotemark[1]\textsuperscript{\rm ~~1}~~~Shuoxi Zhang\textsuperscript{\rm 1}~~~John E. Hopcroft\textsuperscript{\rm 2} \\
\textsuperscript{\rm 1} Huazhong University of Science and Technology \hspace{0.3cm} \textsuperscript{\rm 2} Cornell University \hspace{0.3cm} 
}

\begin{document}
\maketitle
\begin{abstract}
Class-Incremental Learning (CIL) is a critical capability for real-world applications, enabling learning systems to adapt to new tasks while retaining knowledge from previous ones. 
Recent advancements in pre-trained models (PTMs) have significantly advanced the field of CIL, demonstrating superior performance over traditional methods. However, understanding how features evolve and are distributed across incremental tasks remains an open challenge. 
In this paper, we propose  a novel approach to modeling feature evolution in PTM-based CIL through the lens of neural collapse (NC),
a striking phenomenon observed in the final phase of training, which leads to a well-separated, equiangular feature space.  
We explore the connection between NC and CIL effectiveness, showing  that aligning feature distributions with the NC geometry enhances the ability to  capture the dynamic behavior of continual learning. Based on this insight, we introduce {\bf N}eural {\bf C}ollapse-inspired {\bf P}re-{\bf T}rained {\bf M}odel-based {\bf CIL} (\name), a method that dynamically adjusts the feature space to conform to the elegant NC structure, thereby enhancing the continual learning process. Extensive experiments demonstrate that \name outperforms state-of-the-art methods across four benchmark datasets. Notably, when initialized with ViT-B/16-IN1K, \name surpasses the runner-up method by 6.73\% on VTAB, 1.25\% on CIFAR-100, and 2.5\% on OmniBenchmark.




\end{abstract}    
\section{Introduction}
\label{sec:intro}

In recent decades, deep learning has made remarkable strides in the field of computer vision, leading to significant advancements in performance across various downstream tasks, including image classification~\cite{he2016deep,he2016identity,hu2018squeeze, dosovitskiy2020image}, object recognition~\cite{girshick2015fast,lin2017feature,chen2019mmdetection}, and semantic segmentation~\cite{zhao2017pyramid, mmseg2020, zhong2023understanding}, among others. However, most existing models 
are designed to operate in an offline training setup, limiting their ability to adapt once deployed. In real-world scenarios, data arrives in a continuous stream, with new categories constantly emerging.  Consequently, deployed  models are prone to performance degradation and obsolescence,  as they fail to accommodate the evolving data distribution.

To mitigate the resource burden associated 
of constant retraining, a widely adopted approach is {C}lass-{I}ncremental {L}earning (CIL), which allows models to incrementally update their knowledge with data from new categories.   However, CIL is often plagued  by the challenge   of \emph{catastrophic forgetting}, wherein the acquisition of new knowledge leads to a significant decline in performance on previously learned tasks.

Conventionally, CIL methods  train models {\sc from scratch} with randomly initialized weights.  
However, with the rise  of {\sc Pre-Trained Models}~ (PTMs) ~\cite{dosovitskiy2020image}, combining PTMs with CIL has emerged as a promising solution.  
PTMs, with their powerful generalization capabilities, have shown substantial improvements over methods trained from scratch, making them a valuable asset for CIL. 

Inspired by the efficient tuning of large language models~\cite{lester2021power} , recent PTM-based CIL methods have adopted prompt-based mechanisms~\cite{wang2022learning,wang2022dualprompt,smith2023coda}. These methods freeze the pre-trained backbone and append a small set of additional tokens (\ie, prompts)  to the input, which are optimized to adapt to incremental tasks, typically by selecting and refining task-specific prompts from a prompt pool to improve model adaptability. 
While these approaches show promise, the underlying mechanism of prompt optimization remains inadequately understood, and the overlap of prompts between tasks 
can lead to conflicts, causing catastrophic forgetting.

Another line of research, feature-based PTM-CIL~\cite{zhou2024expandable,zhou2024revisiting,mcdonnell2024premonition,zhang2023slca}, leverages the strong semantic representations of  PTMs to guide continual learning. These approaches assume that PTMs capture rich feature representations, enabling effective generalization to continual learning tasks. Typically, they often rely on  prototype-based classification, where each instance is assigned to the class whose centroid is most similar. However, due to the inherent feature drift in continual learning, it remains unclear whether this na\"{i}ve paradigm results in an optimal classifier space. Furthermore, the question of  \emph{how feature distributions  evolve across tasks in continual learning scenarios}  has not been adequately explored . We argue that understanding these dynamics is essential for  fully leveraging the semantic power of pre-trained models in CIL. 

In this paper, we investigate how feature distributions should  evolve in PTM-based CIL. Specifically, we model this evolution using neural collapse (NC)~\cite{papyan2020prevalence}.  
NC describes a geometric structure that emerges in the final phase of training, where features collapse to their respective class prototypes, 
resulting in a well-separated, simplex equiangular tight frame (ETF).  
We begin with an empirical analysis of the relationship between NC and the effectiveness of CIL methods. Observing the benefits of NC for continual learning, we propose Neural Collapse-inspired Pre-trained Model-based Class-Incremental Learning (\name). In this approach, we leverage the semantic features of pre-trained models to fit an ETF-based NC structure. Rather than using a fixed ETF classifier, we design a {\sc dynamic ETF classifier} that can incorporate new classes and adjust to preserve the NC structure. This adjustment is achieved through an {\sc ETF alignment}. Additionally, we introduce a {\sc pull-and-push loss}, which drives features toward their classifier prototypes while maximizing class separation, thus improving the decision boundary.

In our extensive experimental evaluation, \name achieves state-of-the-art performance on most datasets. Notably, when initialized with ViT-B/16-IN1K, \name surpasses the runner-up method by 6.73\% on VTAB, 1.25\% on CIFAR-100, and 2.5\% on OmniBenchmark. 
Furthermore, when compared to the upper bound of joint learning, \name achieves results that are within 1.5\% (CIFAR-100) and 0.52\% (OmniBenchmark) of the upper bound, showcasing its potential to approach the theoretical limit of continual learning. 

Our main contributions can be summarized as follows:
\begin{itemize}
    \item We model feature evolution in PTM-based CIL using the NC structure
, providing a new perspective on the dynamics of continual learning. 

    \item We introduce a dynamic ETF classifier, ETF alignment, and PAP loss, which together constitute the core contributions of our method, enabling the dynamic maintenance of the NC structure in continual learning tasks.

    \item Our experimental results highlight the superiority of \name and underscore the potential of neural collapse as a promising framework for advancing continual learning.
    
\end{itemize}


\section{Related Work}

\begin{figure*} [t]
\centering
\subfloat[Fine-tune ViTs]{
    \includegraphics[scale=0.37]{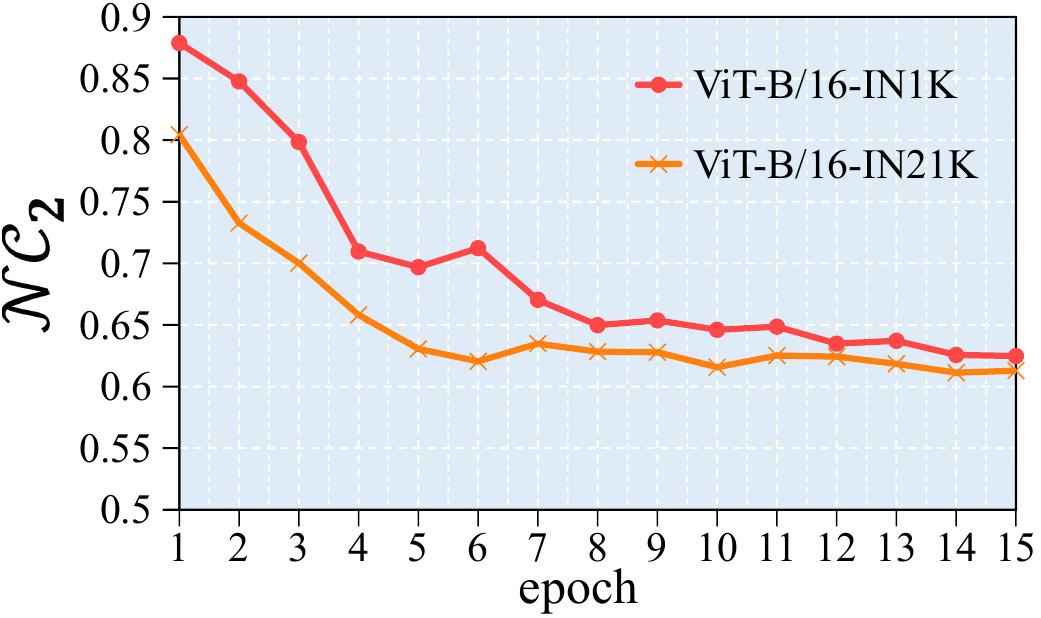}
    \label{fig:sub1}
    }
\quad\quad
\subfloat[PTM-CIL]{
    \includegraphics[scale=0.37]{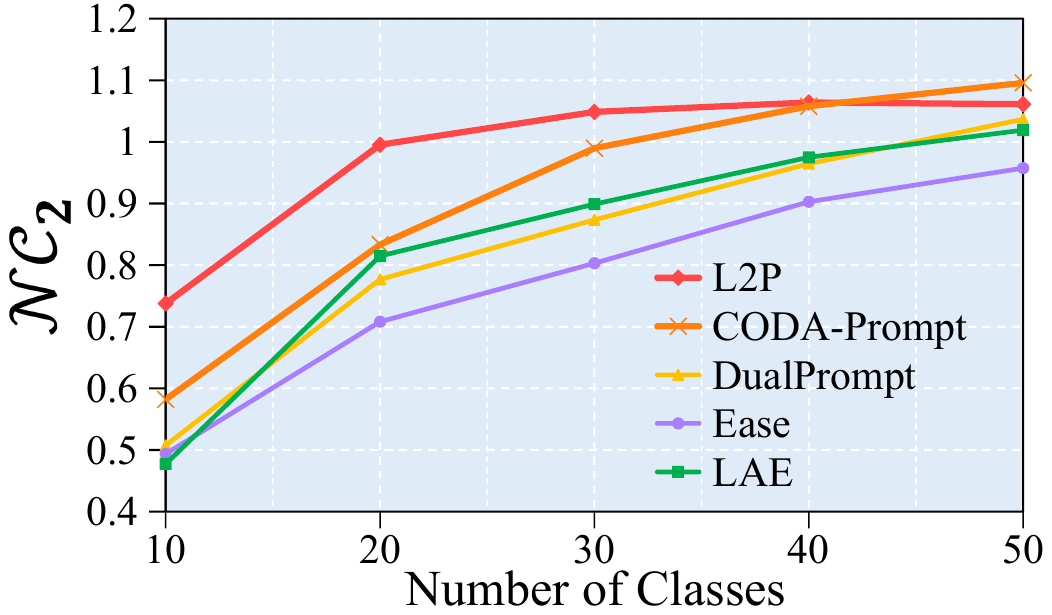}
    \label{fig:sub2}
    }

\caption{(a): The $\mathcal{NC}_\mathbf{2}$ metric with respect to the number of epochs during fine-tuning for ViT-B/16-IN1K and ViT-B/16-IN21K. (b): The $\mathcal{NC}_\mathbf{2}$ values of different PTM-CIL methods on the VTAB dataset during the incremental learning phase, where VTAB is divided into five incremental learning tasks, each containing 10 classes.}
\label{fig: Investigating} 
\end{figure*}
\subsection{Class-Incremental Learning}
Class-incremental learning requires a model to continuously acquire new class semantics while retaining knowledge from previous tasks. One group of approaches~\cite{serra2018overcoming,rusu2016progressive,wang2022coscl,yang2022continual} involves progressively expanding the network architecture, with a dedicated parameter subspace allocated for each new task to mitigate forgetting of prior knowledge. Another line of research~\cite{prabhu2020gdumb,buzzega2020dark,wang2022memory,wang2021ordisco,wang2021triple} focuses on storing raw data from previous tasks, which can be combined with new data during the training of subsequent tasks to mitigate hyperplane deviation from the original feature space.  Regularization-based approaches primarily propose applying penalty terms to regularize training parameters~\cite{chaudhry2018riemannian,kirkpatrick2017overcoming,zenke2017continual,lee2017overcoming} or to distill knowledge~\cite{hinton2015distilling} from the old task model~\cite{li2017learning,rebuffi2017icarl,wu2019large,zhang2020class,hou2019learning,kang2022class,douillard2020podnet,liu2022model}.

\subsection{Pre-trained Model-based Class-Incremental Learning}
Pre-trained models generally exhibit powerful semantic representation capabilities. Consequently, with the increasing prominence of pre-trained models, their integration with incremental learning has become a key research focus within the field of class-incremental learning (CIL)~\cite{mcdonnell2024premonition,wang2024hierarchical,zhou2024continual}. In general, approaches for leveraging pre-trained models in incremental learning tasks fall into two main categories. The first category~\cite{wang2022learning, wang2022dualprompt,smith2023coda} optimizes prompts during training, which are added to the input to encode task-specific information, thereby preserving old knowledge. The second research line focuses on utilizing the robust semantic representations of pre-trained models to improve performance in continual learning. For example, Aper~\cite{zhou2024revisiting} leverages the semantic information from pre-trained features and accumulates class means as classifier. Similarly, \cite{mcdonnell2024premonition} employs an online LDA classifier to reduce class-wise correlations, thereby improving separability.  Moreover, \cite{zhou2024expandable} allocates task-specific subspaces to each task.


\subsection{Neural Collapse}

Neural Collapse (NC) describes an elegant mathematical structure that emerges in the terminal phase of deep model training, in which the features of the last layer and the classifier converge to form a simplex equiangular tight frame~\cite{papyan2020prevalence}. Numerous studies have consistently demonstrated the pervasive occurrence of NC across different settings, including models trained with cross-entropy~\cite{wojtowytsch2020emergence, graf2021dissecting,lu2022neural,fang2021exploring,zhu2021geometric,ji2021unconstrained}, mean squared error (MSE)~\cite{mixon2022neural,poggio2020explicit,zhou2022optimization,han2021neural,tirer2022extended}, and even within contrastive learning~\cite{wang2023towards} frameworks. Recent studies have attempted to induce NC in various domains, including imbalanced learning~\cite{yang2022inducing,xie2023neural,thrampoulidis2022imbalance,behnia2023implicit}, semantic segmentation~\cite{zhong2023understanding}, transfer learning~\cite{galanti2021role,li2022principled}, and federated learning~\cite{huang2023neural}. 

\begin{definition}[Simplex ETF]
	\label{Simplex ETF}
	A simplex equiangular tight frame consists of a set of vectors $\mathbf{m}_i \in \mathbb{R}^d$, where $i = 1, 2, \cdots, K$, with $d \ge K-1$, satisfying the following condition:
	\begin{equation}
		\mathbf{M} = \sqrt{\frac{K}{K-1}} \mathbf{U} \left( \mathbf{I}_K - \frac{1}{K} \mathbf{1}_K \mathbf{1}_K^T \right),
	\end{equation}
	where $\mathbf{I}_K \in \mathbb{R}^{K \times K}$ is the identity matrix, $\mathbf{1}_K \in \mathbb{R}^K$ is a vector of all ones, and $\mathbf{U} \in \mathbb{R}^{d \times K}$ is a partial orthogonal matrix, satisfying $\mathbf{U}^T \mathbf{U} = \mathbf{I}_K$.
\end{definition}
For simplicity, let the last-layer feature $\vec{f}(\vec{x}_i^{k})$ of the sample $\vec{x}_i^{k}$ (the $i$-th sample from the $k$-th class) be denoted as $\vec{h}_i^{k}$. The $k$-th {\sc class means} and {\sc global mean} of the features are calculated as follows:$$
\vec{h}^k := \frac{1}{N}\sum_{i=1}^{N} \vec{h}^k_{i}, \qquad
\vec{h}^G := \frac{1}{K}\sum_{k=1}^K \vec{h}^k.
$$ 
It is assumed that each class contains $N$ samples, and $K$ denotes the number of classes. Neural collapse can be quantified using the following metrics\cite{papyan2020prevalence}:

\begin{enumerate}
    \item
    \textbf{NC1: Within-class variability collapse.} $\mathcal{NC}_\mathbf{1}$ describes the relative magnitude of within-class variability, defined as $\mathbf{\Sigma_W} = \frac{1}{NK} \sum_{k=1}^K\sum_{n=1}^N(\vec{h}^k_{i}-\vec{h}^k)(\vec{h}^k_{i}-\vec{h}^k)^\top$, in relation to the total variability. To compute $\mathcal{NC}_\mathbf{1}$, we use the within-class covariance $\mathbf{\Sigma_W}$ and the between-class covariance $\mathbf{\Sigma_B} = \frac{1}{K}\sum_{k=1}^K (\vec{h}^k - \vec{h}^G)(\vec{h}^k - \vec{h}^G)^\top$. The collapse of $\mathcal{NC}_\mathbf{1}$ can be quantified by comparing the magnitude of the between-class covariance $\mathbf{\Sigma_B} \in \mathbb{R}^{d\times{d}}$ to the within-class covariance $\mathbf{\Sigma_W} \in \mathbb{R}^{d\times{d}}$ of the learned features using the following relationship:
    \begin{align}\label{eq:NC1}
    \mathcal{NC}_\mathbf{1}\;:=\;\frac{1}{K}\mathrm{Trace}\left(\mathbf{\Sigma_W}\mathbf{\Sigma_B}^\dagger\right),
    \end{align}
    where $\mathbf{\Sigma_B}^\dagger$ is the pseudoinverse of $\mathbf{\Sigma_B}$, and $\mathrm{Trace}$ is the sum of the diagonal elements of a matrix.
    
    \item
    \textbf{NC2: Convergence of the class mean matrix $H$ to Simplex ETF.} For the class mean matrix $H = [\vec{h}^1 - \vec{h}^G,\cdots,\vec{h}^K - \vec{h}^G]$, we measure its proximity to the Simplex ETF through:

    \begin{equation}
\mathcal{NC}_\mathbf{2} = \left\|\frac{HH^{\top}}{\left\|HH^{\top}\right\|_{F}} - \frac{1}{\sqrt{K-1}}\left( \mathbf{I}_K - \frac{1}{K} \mathbf{1}_K \mathbf{1}_K^T \right)\right\|_{F}.      
    \end{equation} 
    where $\left\|\cdot\right\|_{F}$ is the Frobenius norm.

    \item
    \textbf{NC3: Convergence to self-duality.}
    The within-class means centered by the global mean will align with their corresponding classifier weights, which implies that the classifier weights $W$ will converge to the same simplex ETF:

        \begin{equation}
        \mathcal{NC}_\mathbf{3} = \left\|\frac{HW}{\left\|HW\right\|_{F}} - \frac{1}{\sqrt{K-1}}\left(\boldsymbol{I}_{K}-\frac{1}{K}\boldsymbol{1}_{K}\boldsymbol{1}_{K}^{\top}\right)\right\|_{F}.
        \label{eq:nc3}
    \end{equation}

\end{enumerate}

\section{Preliminaries}
\begin{figure*}[t]
\vspace{-35pt}
\centering
\includegraphics[width=0.83\linewidth]{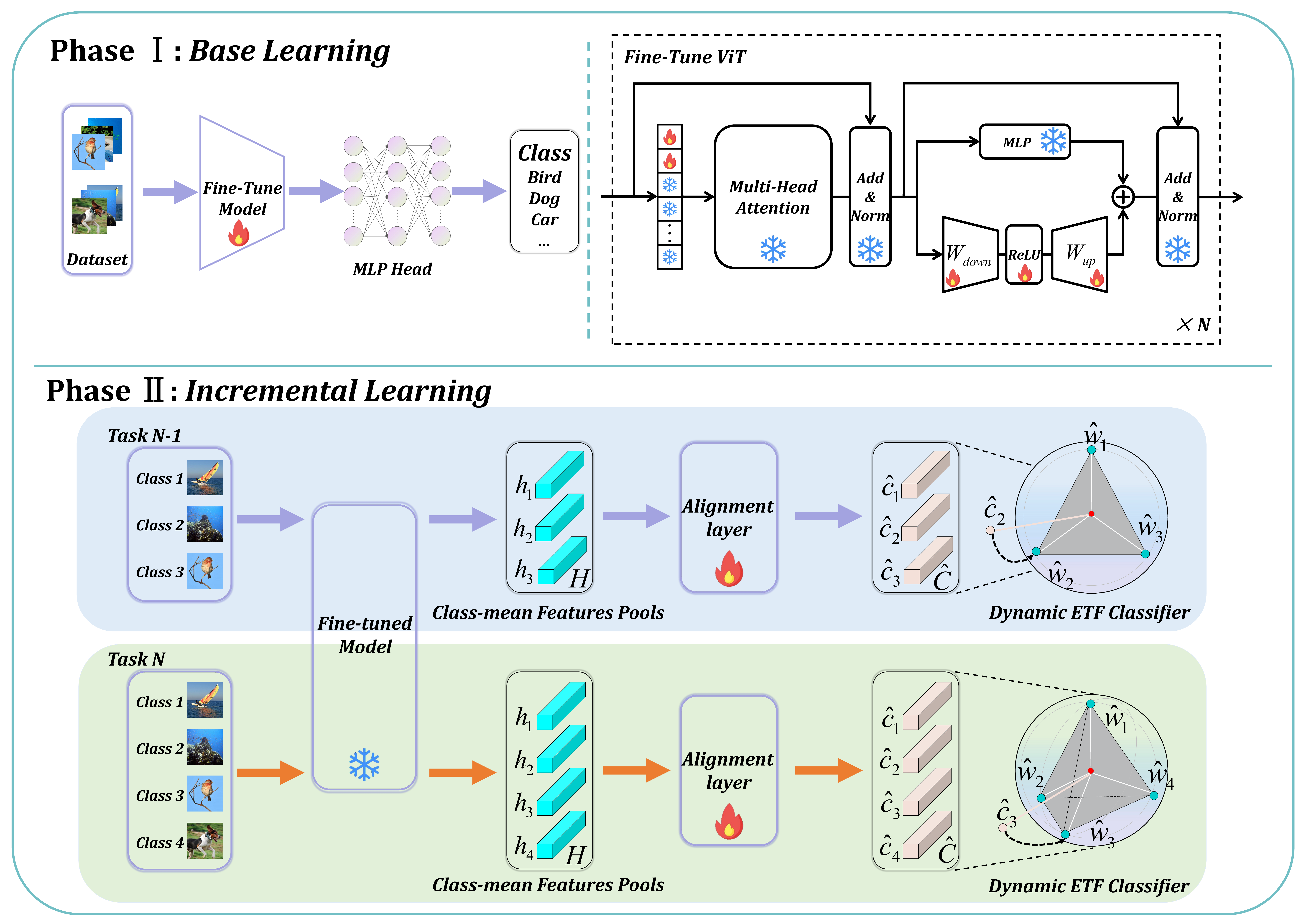} 

\vspace{-5pt}
\caption{Illustration of \name. Phase I: Base Learning. We utilize the initial task dataset \( \mathcal{D}_1 \) to fine-tune the ViT pre-trained model using the VPT Deep~\cite{jia2022visual} and AdaptFormer~\cite{chen2022adaptformer} methods. Phase II: Incremental Learning. We freeze the fine-tuned ViT pre-trained model to extract the class mean features and store them in Class-Mean Features Pools. Subsequently, these class mean features are aligned and mapped to a Dynamic ETF Classifier through an Alignment layer. It is worth noting that the number of vertices in the Dynamic ETF Classifier is adjusted according to the number of currently learned classes. For example, if the total number of learned classes in Task \( N-1 \) is 3, then the number of classifier weight vectors is 3. If the total number of learned classes in Task \( N \) is 4, then the number of classifier weight vectors is 4.}
\label{fig:fig_Framework}
\vspace{-10pt}
\end{figure*}
\subsection{Class-Incremental Learning Setup}
In the scenario of class-incremental learning, a deep model $\Phi(\vec{f}(\cdot))$, composed of a classifier $\Phi(\cdot)$ and a backbone $\vec{f}(\cdot)$, aims to sequentially learn to classify an expanding set of classes so that the model can classify test samples from any task. Specifically, at each incremental step \( t \), the model is presented with a unique set of classes \( \mathcal{C}_t \) and corresponding data $ \mathcal{D}_t = \{(\vec{x}_i, y_i) \mid y_i \in \mathcal{C}_t\} $, where $ \mathcal{C}_t \cap \mathcal{C}_{t'} = \emptyset $ for any $ t \neq t' $. The task is for $\Phi(\vec{f}(\cdot))$ to learn each $ \mathcal{C}_t $ in sequence without revisiting the previous data. The performance of the model at the $T$-th stage is evaluated on all classes seen, $\mathcal{C} = \bigcup_{t=1}^T \mathcal{C}_t$, after each incremental task.
The disjoint nature of classes across steps $t$ introduces the challenge of \emph{catastrophic forgetting}. As new class sets $\{\mathcal{C}_{t+1}, \mathcal{C}_{t+2}, \dots\}$ arrive, the feature representations and decision boundaries are easily modified, causing the model to forget previously learned classes.


\subsection{Investigating Catastrophic Forgetting in PTM-based Incremental Learning} \label{sec:Investigating CF}
 This section investigates the catastrophic forgetting phenomenon in pre-trained model-based class incremental learning. Prior studies\cite{papyan2020traces,zarka2020separation} suggest that deep neural networks progressively enhance linear separability between different classes during training. To validate this hypothesis, we analyze the learning capacity of ViT-B/16-IN1K and ViT-B/16-IN21K pretrained models using the $\mathcal{NC}_\mathbf{2}$ metric (where lower values indicate better linear separability), which measures linear separability through the lens of Simplex Equiangular Tight Frame theory --- a mathematical framework that maximizes inter-class variability for optimal classification\cite{papyan2020prevalence}. We apply the VPT Deep and AdapterFormer methods to fine-tune the pre-trained ViT on the 10-class sampling dataset from VTAB. As shown in \cref{fig:sub1}, enhancing inter-class linear separability constitutes the essential learning mechanism in pre-trained model adaptation.

A critical question arises: \textbf{Does catastrophic forgetting in PTM-CIL primarily manifest as reduced linear separability for previously learned classes?} To answer this, we monitor $\mathcal{NC}_\mathbf{2}$ trajectories across different PTM-CIL methods on the VTAB dataset during incremental learning phases. As shown in \cref{fig:sub2}, sequential incremental learning phases induce significant $\mathcal{NC}_\mathbf{2}$ elevation (i.e., reduced separability) for previous classes. This empirical evidence supports our conclusion: \textbf{Catastrophic forgetting in PTM-CIL essentially stems from the attenuation of inter-class linear separability capacity}.

Motivated by the Simplex ETF framework\cite{zhu2021geometric}, which minimizes class interference through equiangular tight frame-based feature distribution, we propose employing Dynamic ETF Classifier to guide PTM features towards optimal linear separability states. This approach maintains maximal inter-class separation while ensuring uniform feature space distribution, effectively combating catastrophic forgetting through geometric regularization.

\section{Methodology}

As discussed earlier, we have confirmed the feasibility of using NC to model PTM-based continual learning. However, in class-incremental learning, the progressive increase in the number of classes across tasks presents a key challenge: {\it adapting the NC structure to accommodate this growth}. In this section, we introduce our proposed method --- \name --- and provide a detailed illustration of the overall training pipeline, the training loss, and the strategy for dynamically updating the ETF classifier.

\subsection{The Overall Pipeline}
The proposed \name operates in two main phases to leverage the strong generalization of pre-trained models for improving class-incremental learning. The overall pipeline is illustrated in \cref{fig:fig_Framework}.

During the base learning phase, the pre-trained model (ViT, as used in our approach) is fine-tuned on the initial task, adapting it to the target domain while preserving foundational features to support future tasks. Specifically, we employ AdaptFormer~\cite{chen2022adaptformer} and VPT Deep~\cite{jia2022visual} to fine-tune the pre-trained ViT, which inserts trainable low-rank matrices and learnable prompts into the layers of the pre-trained ViT, enabling only these parameters to be updated while keeping the main model weights frozen to swiftly adapting to the new task.  

In the incremental learning phase, the fine-tuned model is frozen, and only the added alignment layer is utilized to maintain and evolve the Neural Collapse (NC) structure. This alignment layer takes class means as input to generate the evolving NC structure, ensuring that the principles of neural collapse are preserved during task updates (e.g., transitioning from 3-ETF to 4-ETF, as shown in \cref{fig:fig_Framework}). 
Finally, the updated ETF structure can serve as the classifier ${\hat{\vec{W}}_{\mathrm{ETF}}} = [\hat{\vec{w}}_{1},\cdots,\hat{\vec{w}}_{K_t}]\in\mathbb{R}^{d\times K_t}$, where $\hat{\vec{w}}_{k}$ is the fixed weight vector for class $k$ in ${\hat{\vec{W}}_{\mathrm{ETF}}}$,  and $K_t$ is the total number of classes at current $t$-th task. During inference, the model computes the maximum similarity between the feature of the instance and $K_t$ classifier vectors to predict the correct category. 
\begin{equation}
\begin{aligned}
y_{\mathrm{test}}=\arg\max_{y^{\prime}\in\{1,...,K_t\}}s_{y^{\prime}},\\ 
\quad s_{y}=\frac{\vec{f}(\vec{x}_i)^{\top}\hat{\vec{w}}_{k}}{\|\vec{f}(\vec{x}_i)\|\cdot\|\hat{\vec{w}}_{k}\|}.
\end{aligned}
\label{eq:etf_classifier}
\end{equation}

\subsection{NC-inspired Function}
\noindent {\bf Dynamic ETF Classifier.} In \cite{zhu2021geometric}'s method, the Simplex ETF classifier is fixed and non-learnable, requiring prior knowledge of the entire label space across future tasks --- a limitation that is impractical for real-world incremental learning scenarios. Dynamic ETF Classifier we propose can increase the number of category prototypes as the number of learned classes grows, thereby better adapting to changes in the real world. Moreover, incorporating the Dynamic ETF Classifier ensures that the class mean features output by the model better align with the $\mathcal{NC}_\mathbf{2}$ characteristics. The formula of the Dynamic ETF Classifier can be constructed as:
\begin{equation} {\hat{\vec{W}}_{\mathrm{ETF}}} = \sqrt{\frac{K_t}{K_t-1}} \mathbf{U}_t \left( \mathbf{I}_{K_t} - \frac{1}{K_t} \mathbf{1}_{K_t}\mathbf{1}_{K_t}^\top \right) \end{equation}
$\mathbf{U}_t \in \mathbb{R}^{d \times K_t}$ is a dynamically expanded orthogonal matrix satisfying: $\mathbf{U}_t^\top\mathbf{U}_t = \mathbf{I}_{K_t}$.
\begin{figure}
    \centering
    \includegraphics[width=0.95\linewidth]{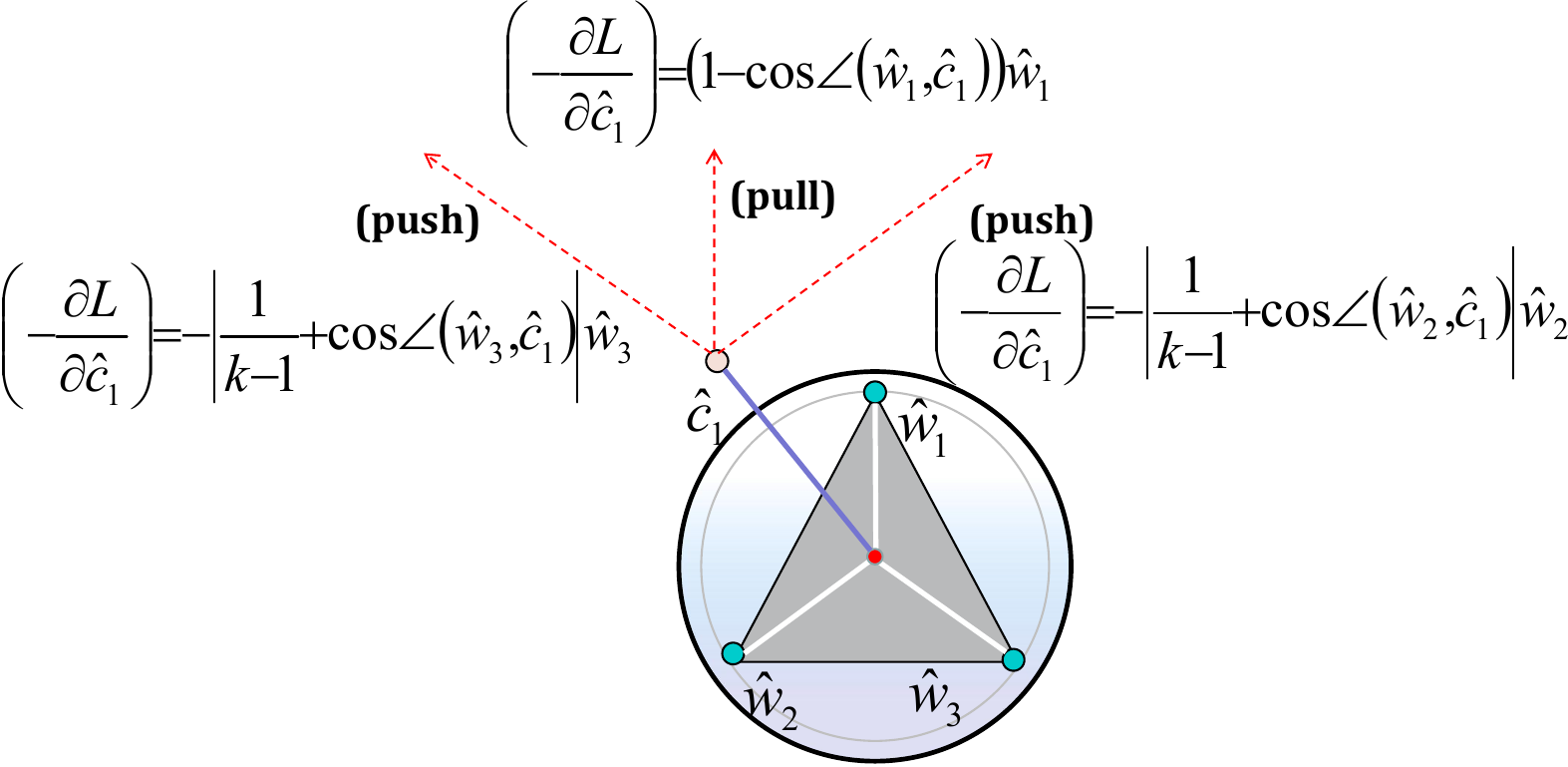}
    \caption{Description of the gradient of PAP Loss with respect to $\hat{c}_1$. For PAP Loss, the gradient ($-\frac{\partial L}{\partial\hat{c}_1}$) pulls $\hat{c}_{1}$ towards $\hat{\vec{w}}_{i}$, while the gradients from ($-\frac{\partial L}{\partial\hat{c}_1}$) and ($-\frac{\partial L}{\partial\hat{c}_1}$) act as a repulsive force, pushing $\hat{c}_{1}$ away from $\hat{\vec{w}}_{i}$ (the red dashed line indicates the direction of the gradient).}
    \label{fig:dynamics}
\end{figure}

\noindent {\bf Pull And Push Loss.} During the incremental learning phase, we design the pull and push (PAP) loss to optimize the parameters of the Alignment layer. This loss function consists of two components: push loss and pull loss. The pull loss drives the class means generated by the PTM to align with their corresponding fixed weight vectors, while the push loss propels class means away from dissimilar vectors. The optimization of PAP loss aims to achieve alignment between the class means extracted by the PTM and the fixed weight vectors in the Dynamic ETF Classifier, thereby maximizing inter-class separability.
\begin{equation}
   \mathcal{L}_\text{PAP}(\hat{\vec{c}}_{k}, {\hat{\vec{W}}_{\mathrm{ETF}}})  =
 \underbrace{( \hat{\vec{w}}_{k}^{T} \hat{\vec{c}}_{k} - 1 )^2}_{pull} + \underbrace{{\sum_{j \neq k}^{K_t} (  \hat{\vec{w}}_{j}^{\top} \hat{\vec{c}}_{k} +\frac{1}{k-1}  )^2}}_{push}  .
   \label{eq: PAP loss}
\end{equation}
where $\hat{\vec{c}}_{k}$ represents the feature vector derived from the class mean $\mathbf{h}^{k}$ through the alignment layer. Notably, the gradient of $\mathcal{L}_\text{PAP}$ with respect to $\hat{\vec{c}}_{k}$ is:·
\begin{equation}
\begin{aligned}
\left(-\frac{\partial\mathcal{L}_{\mathrm{PAP}}}{\partial\hat{\vec{c}}_{k}}\right) & =\underbrace{(1-\cos\angle(\hat{\vec{w}}_k,\hat{\vec{c}}_k))\hat{\vec{w}}_k}_{\mathrm{pull}} \\
 & -\underbrace{\left(\sum_{j\neq k}^{K_t}\left|\cos\angle(\hat{\vec{w}}_j,\hat{\vec{c}}_k)+\frac{1}{k-1}\right|\hat{\vec{w}}_j\right)}_{\mathrm{push}}
\end{aligned}
\label{gradient_pap}
\end{equation} 
As shown in Fig.~\ref{fig:dynamics}, the pull term encourages $\hat{\vec{c}}_{k}$ and $\hat{\vec{w}}_{k}$ to align by reducing the angle between their feature vectors. Specifically, when the cosine similarity $\cos \angle (\hat{\vec{w}}_k, \hat{\vec{c}}_k)$ is low, the gradient magnitude becomes larger, and its direction aligns with $\hat{\vec{w}}_{k}$ (as derived in Eq.~\eqref{gradient_pap}). This directs $\hat{\vec{c}}_{k}$ to be updated towards $\hat{\vec{w}}_{k}$, effectively increasing their similarity. Conversely, the push term penalizes high cosine similarity between $\hat{\vec{c}}_{k}$ and negative weight vectors $\hat{\vec{w}}_j$ ($j \neq k$). When $\cos \angle (\hat{\vec{w}}_j, \hat{\vec{c}}_k)$ is high, the gradient magnitude increases and points away from $\hat{\vec{w}}_j$, driving $\hat{\vec{c}}_{k}$ to orthogonalize with irrelevant classifier weights. It can be concluded that global optimality occurs when $\cos\angle(\hat{\vec{w}}_{k}, \hat{\vec{c}}_{k}) = 1$ and $\cos\angle(\hat{\vec{w}}_{j}, \hat{\vec{c}}_{k}) = -\frac{1}{k-1}$, which theoretically demonstrates that the class mean matrix ultimately forms a simplex ETF, satisfying the $\mathcal{NC}_\mathbf{2}$ property.  

\subsection{ETF Alignment}
   
Considering the feature drift that often occurs in class-incremental learning, using class means to represent the ETF classifier in the current task would lead to significant disruption of the ETF structure. This is where the {\sc alignment layer} plays a crucial role. The alignment
layer is an MLP block following Hersche~\cite{hersche2022constrained}.  Thanks to the fine-tuning capability of the alignment layer, it helps preserve the orthogonality and separability of previously learned classes during the dynamic update of the ETF. Specifically, in Task \( t \), the class means extracted for each class are stored in the {\sc Class-mean Features Pools}, forming the matrix \( \vec{H} \in \mathbb{R}^{d\times K_t} \), where \( K_t\) is the total number of classes at the current $t$-th task. The classifier prototypes are initialized as \( \hat{\vec{W}}_{\mathrm{ETF}} \in \mathbb{R}^{d\times K_t} \). It is important to note that, to prevent the alignment layer parameters from task 
$t-1$ from negatively impacting the alignment for task $t$, we initialize a new set of alignment layer parameters for each task to ensure proper alignment during training. The total loss in our framework can be illustrated as:
\begin{equation}
    \mathcal{L}_\text{total} = \mathcal{L}_\text{CE} + \mathcal{L}_\text{PAP}.
    \label{eq:total_loss}
\end{equation}
where $\mathcal{L}_\text{CE}$ denotes the cross-entropy loss.
\section{Experiments}

In this section, we detail the experimental setup and evaluation metrics, followed by comparisons with state-of-the-art algorithms on five benchmark datasets. Additionally, we conduct an ablation study to assess the robustness of our proposed method and analyze the impact of $\mathcal{NC}$ metrics on model performance. Visualizations are provided in the supplementary material to demonstrate the effectiveness of the model.

\begin{table*}[!t]
    \centering
    \footnotesize
    \begin{tabular}{>{\centering\arraybackslash}p{4cm}| c| c| c| c|c}
        \toprule
        \multirow{2}{*}{Method} & \multirow{2}{*}{Pre-trained} & \multirow{2}{*}{CIFAR-100} & \multirow{2}{*}{CUB-200} & \multirow{2}{*}{VTAB} & \multirow{2}{*}{OmniBenchmark} \\
                                &                               &                             &                            &                                                       &                         \\
        \midrule
        L2P \cite{wang2022learning} \hfill CVPR2022 & IN21K & 87.88 & 77.45 & 71.17 & 74.45 \\ 
        DualPrompt \cite{wang2022dualprompt} \hfill ECCV2022 & IN21K & 89.44 & 84.61 & 88.37 & 76.40 \\ 
        CODA-Prompt \cite{smith2023coda} \hfill CVPR2023 & IN21K & 91.18 & 85.11 & 88.15 & 73.37 \\ 
        
        LAE \cite{gao2023unified} \hfill ICCV2023 & IN21K & 90.34 & 83.40 & 88.92 & 73.09 \\ 
        Aper \cite{zhou2024revisiting} \hfill IJCV2024 & IN21K & 92.25 & 89.10& 90.12 & 77.30 \\ 
        Ease \cite{zhou2024expandable} \hfill CVPR2024 & IN21K & 92.08 & 90.12 & 90.41 & 75.77 \\ 

        \rowcolor{black!15}
        \name(Ours) & IN21K & \textbf{93.51} & \textbf{91.22} & \textbf{92.96} & \textbf{80.51} \\ 
        \bottomrule
    \end{tabular}

    \caption{Average Top-1 accuracy comparison on four datasets using \textbf{ViT-B/16-IN21K} as the backbone. The best performance is highlighted in bold.}
    \label{table:ALL}
\end{table*}

\begin{table*}[h!]
    \centering
    \resizebox{0.78\textwidth}{!}{
    \begin{tabular}{>{\centering\arraybackslash}p{4cm}| c |c |c|c| c}
        \toprule
        \multirow{2}{*}{Method} & \multirow{2}{*}{Pre-trained} & \multirow{2}{*}{CIFAR-100} & \multirow{2}{*}{CUB-200} & \multirow{2}{*}{VTAB} & \multirow{2}{*}{OmniBenchmark} \\
                                &                               &                             &                            &                             &                                                  \\
        \midrule
        Joint & IN21K & 90.83 & 86.43 & 91.58 & 74.67 \\ 
        \rowcolor{black!15}
        \name (Ours) & IN21K & 89.33 & 85.41 & 86.94 & 74.15 \\ 
        \bottomrule
    \end{tabular}}


    \caption{\name Last Performance Comparison with the Upper Bound for CIL. \enquote{Joint} refers to fine-tuning the pre-trained model using all classes of the dataset, which represents the upper bound in the continual learning paradigm. }
    \label{table:Joint+NCPM_CIL}
\end{table*}
\begin{figure*} [t]
	\centering
	\begin{subfigure}{0.33\linewidth}
		\includegraphics[width=1\columnwidth]{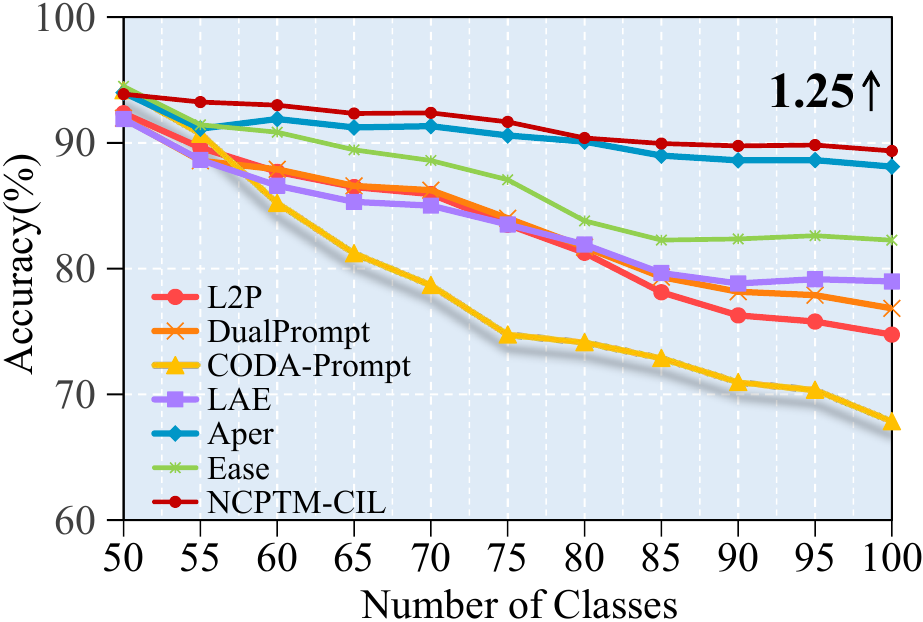}
		\caption{CIFAR-100 B50 Inc5}
		\label{fig:benchmark-cifar100}
	\end{subfigure}
	\hfill
	\begin{subfigure}{0.33\linewidth}
		\includegraphics[width=1\linewidth]{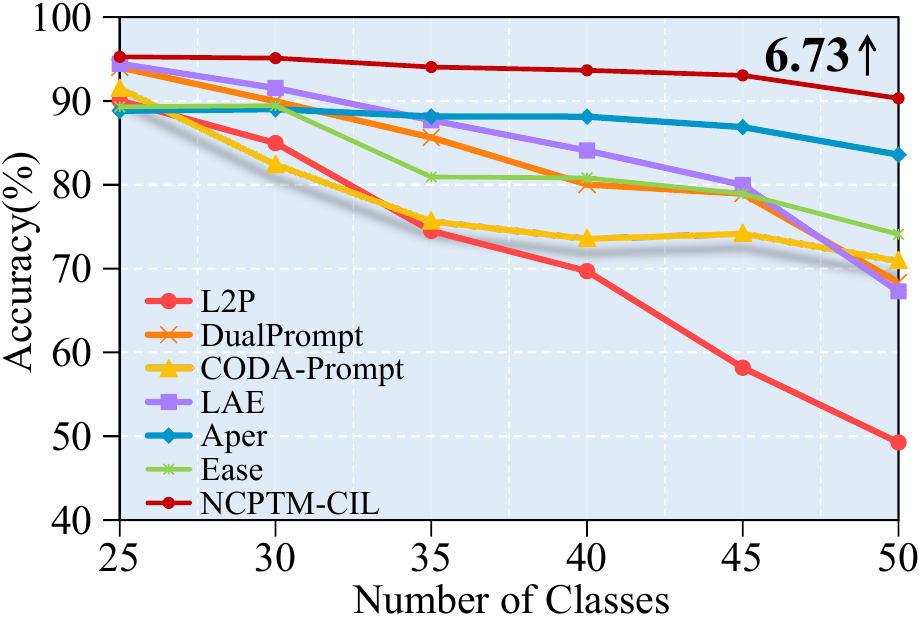}
		\caption{VTAB B25 Inc5}
		\label{fig:benchmark-cub200}
	\end{subfigure}
	\hfill
	\begin{subfigure}{0.33\linewidth}
		\includegraphics[width=1\linewidth]{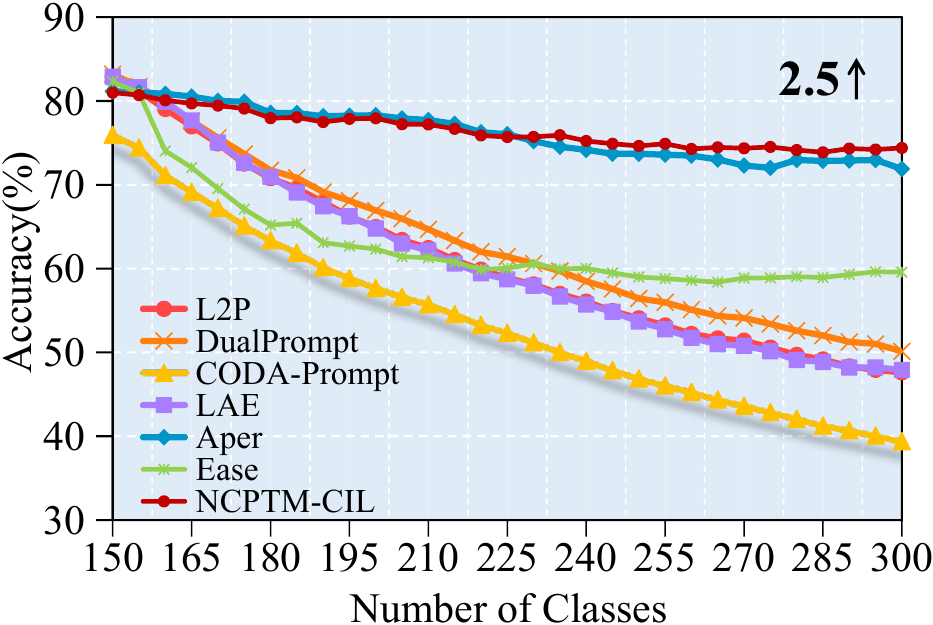}
		\caption{OmniBenchmark B150 Inc5}
		\label{fig:benchmark-omniB}
	\end{subfigure}
	\\
	
	\vspace{-3mm}
	\caption{All methods are initialized with \textbf{ViT-B/16-IN1K}. We annotate the relative improvement of \name over the runner-up method with numerical values at the final incremental stage. }
	\vspace{-3mm}
	\label{fig:IN1K}
\end{figure*}

\subsection{Implementation Details}

\noindent {\bf Dataset:} Since pre-trained models often retain extensive knowledge from upstream tasks, following~\cite{wang2022learning,zhou2024revisiting}, we evaluate performance on CIFAR-100~\cite{krizhevsky2009learning}, CUB-200~\cite{wah2011caltech}, VTAB~\cite{zhai2019large}, and OmniBenchmark~\cite{zhang2022benchmarking}. These datasets include standard class-incremental learning benchmarks as well as out-of-distribution datasets with significant domain gaps relative to ImageNet (the dataset used for pre-training). CIFAR-100 contains 100 classes, VTAB includes 50 classes, CUB-200 has 200 classes, and OmniBenchmark encompasses 300 classes. Additional details are provided in the supplementary material.

\noindent {\bf Training details:} Following ~\cite{zhou2024revisiting}, we consider two representative models, \ie, ViT-B/16-IN21K and ViT-B/16-IN1K as pre-trained models. The former is pre-trained on ImageNet-21K, and the latter is further fine-tuned on ImageNet-1K. In the base learning phase, we use the SGD optimizer to fine-tune the pre-trained model with a batch size of 48 for 20 epochs of training. The learning rate starts at 0.01, weight decay is set to 0.0005, momentum is 0.9, and cosine annealing scheduling is used. In the incremental learning phase, we use the SGD optimizer to train {\sc Alignment layer} and employ cosine annealing scheduling. Typically, we perform 40 epochs of training, but in some experiments, this is reduced to fewer epochs if significant overfitting is observed.

\noindent {\bf Evaluation metrics \& Comparison methods:} Following~\cite{rebuffi2017icarl}, we use $\mathcal{A}_{t}$ to represent the average accuracy across all classes after the $t$-th stage. We use the accuracy after learning the last task $\mathcal{A}_{T}$ and the average accuracy across all tasks $\mathcal{\bar{A }}=\frac{1}{T}\sum_{t=1}^T\mathcal{A}_t$ as evaluation metrics. We compare our method with state-of-the-art PTM-based CIL approaches, including L2P~\cite{wang2022learning}, DualPrompt~\cite{wang2022dualprompt}, CODA-Prompt~\cite{smith2023coda}, LAE~\cite{gao2023unified}, Aper~\cite{zhou2024revisiting}, and EASE~\cite{zhou2024expandable}. For fairness, all methods are evaluated independently using ViT-B/16 as the pre-trained model.

\subsection{Main Results}

We compare \name with other state-of-the-art methods on five standard datasets. Specifically, CIFAR-100, CUB-200, and OmniBenchmark are divided into 10 incremental learning tasks, while VTAB is divided into 5 incremental learning tasks. From the results in \Cref{table:ALL}, it can be observed that \name demonstrates superior performance over all existing methods. To ensure a fair comparison, we also report the performance trend in incremental learning settings across three datasets, as shown in~\cref{fig:IN1K}. In these experiments, the base learning phase involves training on half of the total classes, with 5 new classes introduced at each incremental step. The trend plots illustrate that as the number of classes increases, the accuracy of \name declines only gradually, maintaining a nearly horizontal trajectory compared to other methods. This phenomenon confirms the effectiveness of our method in mitigating catastrophic forgetting. In particular, \name outperforms the runner-up method by 6.73\% on the VTAB dataset, by 1.25\% on CIFAR-100, and by 2.5\% on OmniBenchmark.

It is well-established that joint learning, which involves integrating old data, represents the upper bound in the continual learning paradigm. We aim to evaluate how closely our approach approximates this theoretical performance limit. Therefore, we conduct a comparative evaluation between our method and joint learning. As shown in Table~\ref{table:Joint+NCPM_CIL}, our method achieves performance that is nearly on par with joint learning.  Specifically, the gap between our method and the incremental learning upper bound is 1.5\%, 1.02\%, and 0.52\% on the CIFAR-100, CUB-200, and OmniBenchmark datasets, respectively. 
\begin{table}[h!]
    \centering
    \footnotesize
    \begin{tabular}{>{\centering\arraybackslash}ccc | cc}
    \toprule
        \multicolumn{3}{c|}{Module}  & \multirow{2}{*}{CIFAR-100}\\
         Dynamic-ETF  & Init-Align & PAP Loss &  \\
    \midrule
       \checkmark&---&---    & 92.32  \\ 
       ---&\checkmark&---       & 92.28  \\ 
       ---&---&\checkmark          & 91.92  \\ 
       \checkmark&\checkmark&---          & 93.25  \\
        \checkmark&---&\checkmark         & 93.07  \\ 
        
       ---&\checkmark&\checkmark & 93.21 \\
       \checkmark&\checkmark&\checkmark & \textbf{93.51} \\
    \bottomrule
    \end{tabular}
    \caption{Ablation study results showing the average Top-1 accuracy after continual learning of all classes on the CIFAR-100 dataset. \enquote{Dynamic-ETF} refers to the dynamic ETF classifier, while \enquote{Init-Align} denotes the Initial Alignment Layer.}
    \label{table:Ablation}
\end{table}
\subsection{Ablation Study}

To evaluate the contribution of each component in \name, we conduct an ablation study by removing each module individually. Table~\ref{table:Ablation} shows that removing the Dynamic ETF Classifier results in a noticeable performance drop (from 93.51\% to 93.21\%), indicating its importance in maintaining the equiangular feature structure aligned with neural collapse. The absence of the Initial Alignment Layer reduces accuracy to 93.07\%, highlighting its role in mitigating the negative influence of alignment parameters from previous tasks. This layer ensures that features are optimally aligned for each task, enhancing task-specific adaptability. Excluding PAP Loss leads to a 0.26\% drop, suggesting that it enhances feature separability across classes. These results validate the necessity of each component in shaping a robust feature space for incremental learning.

\begin{table}[h!]
    \centering
    \footnotesize
    \begin{tabular}{>{\centering\arraybackslash}p{5cm} |c| c}
    \toprule
        Fine-Tune Method & $\mathcal{NC}_\mathbf{1}$ & CIFAR-100 \\
    \midrule
        None                  & 0.253 & 88.436  \\ 
        VPT shallow          & 0.252 & 88.518  \\ 
        AdaptFormer           & 0.132 & 92.231  \\ 
        Lora                  & 0.107 & 93.027  \\
        VPT Deep             & 0.101 & 93.115  \\ 
        \rowcolor{black!15}
        AdaptFormer+VPT Deep(\name) & \textbf{0.095} & \textbf{93.513} \\
    \bottomrule
    \end{tabular}
    \caption{Comparison of fine-tuning methods on CIFAR-100. We report the $\mathcal{NC}_\mathbf{1}$ (initial task) and the average top-1 accuracy for each method.}
    \label{table:NC1}
\end{table}
\begin{table}[h!]
    \centering
    \footnotesize
    \begin{tabular}{c | c| c|c }
        \toprule
        \multirow{2}{*}{Method}  & \multirow{2}{*}{CIFAR-100}  & \multirow{2}{*}{$\mathcal{NC}_\mathbf{2}$} & \multirow{2}{*}{$\mathcal{NC}_\mathbf{3}$}  \\
                                                               &                         &                                            &                                        \\
        \midrule
        \rowcolor{black!15}
        \name                                       & \textbf{93.51}    & \textbf{ 0.217}  & \textbf{0.106}    \\ 
        w/o. Dynamic ETF Classifier                    & 93.21   & 0.259        & 0.138   \\ 
        w/o. Initial Alignment Layer             & 93.07   & 0.471        &0.296    \\ 
        w/o. PAP Loss         & 93.25    &0.235     & 0.125  \\ 

        \bottomrule
    \end{tabular}
    \caption{Ablation study results showing the average Top-1 accuracy after continual learning of all classes, along with the average $\mathcal{NC}_\mathbf{2}$ and $\mathcal{NC}_\mathbf{3}$ scores obtained from experiments on the CIFAR-100 dataset.}
    \label{table:Ablation_nc}
    \vspace{-15pt}
\end{table}
\subsection{Rethinking Incremental Learning} 

Given the variety of fine-tuning methods for large models, selecting the most suitable method for different datasets can be both time-intensive and computationally costly. Therefore, we aim to identify an efficient selection strategy for fine-tuning methods from the perspective of NC. Experimentally, we examine the relationship between $\mathcal{NC}_\mathbf{1}$ values in the base learning stage and the final continual learning performance, as shown in \cref{table:NC1}. Clearly, when the NC values in phase 1 are better (\ie, lower $\mathcal{NC}_\mathbf{1}$, indicating better NC), the final continual learning results also improve. This confirms the effectiveness of using NC to model the continual learning distribution and suggests that calculating NC based on first-stage features is sufficient for selecting a suitable fine-tuning method, thereby significantly reducing the time cost of this selection process. Specific details of the fine-tuning methods we used are provided in the supplementary material.

\begin{figure}
    \centering
    \includegraphics[width=0.90\linewidth]{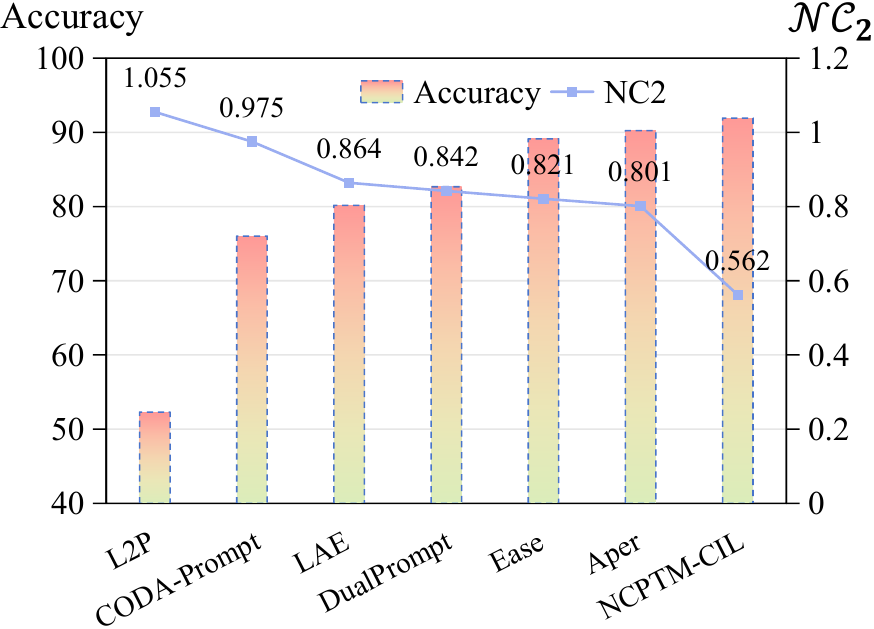}
    \caption{The relationship between the performance of existing PTM-based CIL approaches and their feature's $\mathcal{NC}_\mathbf{2}$. }
    \label{fig:acc_with_nc2}
\end{figure}

We are also interested in how $\mathcal{NC}_\mathbf{2\sim3}$ impacts continual learning, so let us revisit~\cref{table:Ablation_nc}
. As shown in~\cref{table:Ablation_nc}, performance declines as key components are removed, and the corresponding $\mathcal{NC}_\mathbf{2}$ and $\mathcal{NC}_\mathbf{3}$ values increase. Note that lower values of $\mathcal{NC}_\mathbf{2}$ and $\mathcal{NC}_\mathbf{3}$ indicate more pronounced NC. Thus, we conclude that the properties of $\mathcal{NC}_\mathbf{2}$ and $\mathcal{NC}_\mathbf{3}$ contribute to a more discriminative continual learning distribution. This effect may be due to the promotion of a self-dual ETF structure formed by the feature means and classifier weights, resulting in a clearer decision boundary. This aligns with the objective of our PAP loss, further validating the effectiveness of our approach.

To validate if PTM-CIL's effectiveness stems from the $\mathcal{NC}_\mathbf{2}$ property, we assess PTM-CIL accuracy on VTAB and computed features $\mathcal{NC}_\mathbf{2}$. All implementations use \textbf{ViT-B/16-IN21K} with VTAB split into 10 incremental tasks. As illustrated in~\cref{fig:acc_with_nc2}, the average task accuracy of PTM-CIL exhibits a significant correlation with $\mathcal{NC}_\mathbf{2}$. This empirical evidence substantiates that our proposed methodology enhances the linear separability of model-generated features, thereby supporting the hypothesis regarding the critical role of Neural Collapse guidance in facilitating feature evolution during continual learning processes.

\section{Conclusion}

In this paper, we revisited Class-Incremental Learning (CIL) in the context of pre-trained models (PTMs) and investigated feature evolution through the lens of Neural Collapse (NC) theory. Our analysis revealed that NC, a phenomenon in which 
class features align equiangularly at the final stage of training, offers a powerful framework for understanding feature evolution across incremental tasks.     
Building on this insight, we introduced {\bf N}eural {\bf C}ollapse-inspired {\bf P}re-{\bf T}rained {\bf M}odel-based {\bf CIL} (\name), a method designed to maintain a structured and resilient feature space across tasks.  
Key components of our approach include the {\sc Dynamic ETF classifier}, {\sc ETF alignment}, and {\sc PAP loss}, which together maintain a structured and resilient feature space across tasks, thereby effectively mitigating catastrophic forgetting. Extensive experiments validated the effectiveness of our approach, achieving superior performance across multiple benchmarks. These results underscore the efficacy  of our NC-inspired design in 
enabling models to adapt to new classes without forgetting prior ones, marking a significant step forward in leveraging pre-trained models for continual learning.

\newpage
{
    \small
    \bibliographystyle{ieeenat_fullname}
    \bibliography{main}
}

\clearpage
\setcounter{page}{1}
\maketitlesupplementary

\setcounter{section}{0}
\setcounter{equation}{0}

\section{Experimental Settings}

In this section, we provide an overview of the datasets employed in our study:

\noindent {\bf CIFAR-100:} This dataset comprises 60,000 images spanning 100 classes, with 50,000 images designated for training and 10,000 for testing. Each class includes 100 images.

\noindent {\bf CUB-200:} Widely used for fine-grained visual classification tasks, this dataset contains 11,788 images across 200 bird species. The training set has 9,430 images, while the test set includes 2,358 images.

\noindent {\bf OmniBenchmark:} This dataset contains 89,697 training images and 5,985 testing images across 300 diverse classes. Due to the large domain gaps between classes, OmniBenchmark poses a greater challenge for incremental learning compared to other datasets.

\noindent {\bf VTAB:} This dataset supports cross-domain incremental learning across five domains. It consists of 1,796 training images and 8,619 testing images covering 50 classes.


\section{Fine-tuning methods}

We experimented with four fine-tuning methods for the ViT-B/16 pretrained model: LoRA~\cite{zhu2024melo}, AdaptFormer~\cite{chen2022adaptformer}, VPT Deep, and VPT Shallow~\cite{jia2022visual}.

\noindent {\bf LoRA:} The core idea of LoRA~\cite{hu2021lora} is to freeze the weights of the pretrained model and decompose its weight matrices into low-rank matrices, thus reducing the number of trainable parameters and improving fine-tuning efficiency. Following the approach in ~\cite{zhu2024melo}, we inject LoRA weights into the query projection matrix $W_{Q}$ and value projection matrix $W_{V}$ of each self-attention layer in ViT. The formulation can be expressed as: 
\begin{equation}
y=W_0x+BAx
    \label{eq:lora}
\end{equation}
where $x \in \mathbb{R}^{k\times d}$ and $x \in \mathbb{R}^{k\times d}$ represent the input and output features, respectively. $W_0\in \mathbb{R}^{d\times d}$ denotes the pretrained weights for either $W_{Q}$ or $W_{V}$, while $B\in\mathbb{R}^{d\times r}$ and $A\in\mathbb{R}^{r\times d}$ are the trainable low-rank matrices. We set the rank of the low-rank matrices to $r=8$ significantly smaller than the model dimension $d=768$. By using LoRA, we greatly reduce the number of trainable parameters, enhancing training efficiency.

\noindent {\bf AdaptFormer:} AdaptFormer focuses on fine-tuning the MLP layers of ViT. It introduces two projection matrices: $W^{\mathrm{down}}$ for dimensionality reduction, and $W^{\mathrm{up}}$ to project the reduced feature back to the original dimension. Given an input $x\in\mathbb{R}^{k\times d}$, the AdaptFormer output is computed as:
\begin{equation}
y=\mathrm{MLP}\left(x\right)+\mathrm{ReLU}\left(W^{\mathrm{down}}x\right)W^{\mathrm{up}}
    \label{eq:adaptFormer}
\end{equation}
where $W^{\mathrm{down}}\in\mathbb{R}^{d\times r}$ and $W^{\mathrm{up}}\in\mathbb{R}^{r\times d}$. We set the rank $r=64$ , much smaller than the model dimension $d=768$. During fine-tuning, MLP parameters are frozen, and only the two projection matrices are trained, reducing the computational cost.

\noindent {\bf VPT:} VPT adds learnable parameters $\mathbf{P}\in\mathbb{R}^{p\times d}$ to the image encoding features ${x}\in\mathbb{R}^{k\times d}$, forming the final expanded feature:
\begin{equation}
x^{\prime}=[x;p]
    \label{eq:VPT}
\end{equation}
where $x^{\prime}\in\mathbb{R}^{(k+p)\times d}$ is the input to the subsequent layers of ViT. VPT has two variants: VPT Shallow, which attaches the trainable parameters $\mathbf{P}$ only at the first layer of the pretrained model, and VPT Deep, which attaches $\mathbf{P}$ at every layer of the pretrained model.
\begin{figure}[t]
	\centering

	\begin{subfigure}{0.43\linewidth}
		\includegraphics[width=1\linewidth]{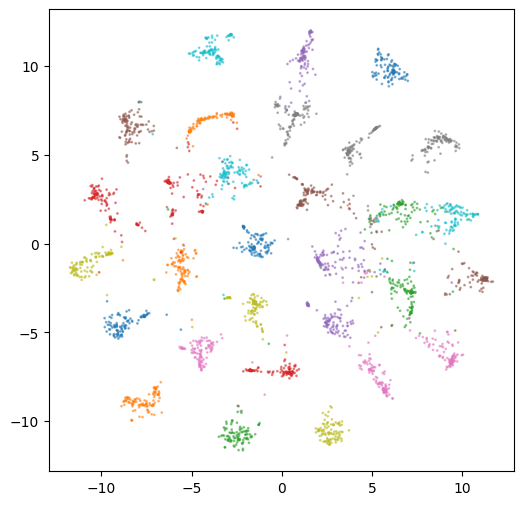}
		\caption{30 classes of Aper}
		\label{fig:Aper30}
	\end{subfigure}
	\hfill
	\begin{subfigure}{0.43\linewidth}
		\includegraphics[width=1\linewidth]{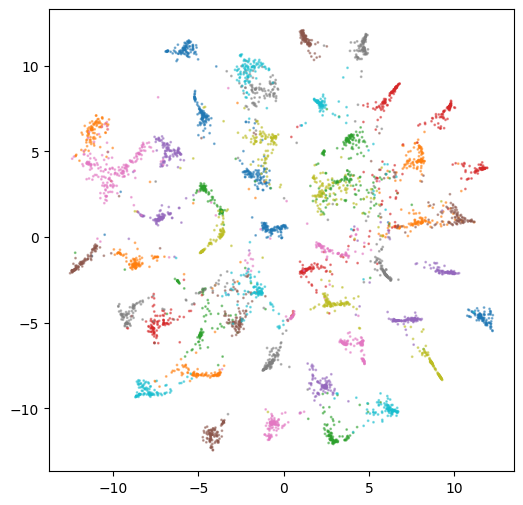}
		\caption{50 classes of Aper}
		\label{fig:Aper50}
	\end{subfigure}
	\\

	\begin{subfigure}{0.43\linewidth}
		\includegraphics[width=1\linewidth]{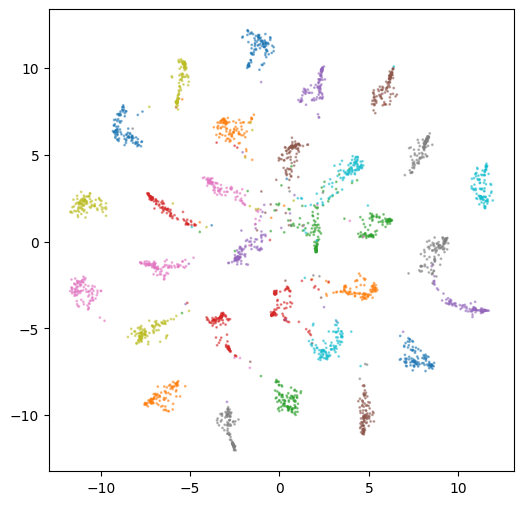}
		\caption{30 classes of \name}
		\label{fig:NCPM_CIL30}
	\end{subfigure}
	\hfill
	\begin{subfigure}{0.43\linewidth}
		\includegraphics[width=1\columnwidth]{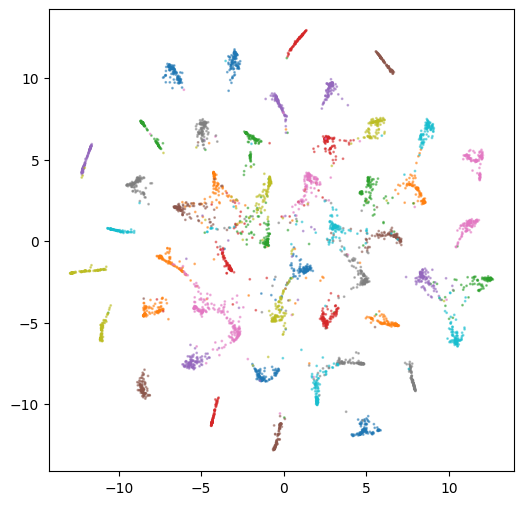}
		\caption{50 classes of \name}
		\label{fig:NCPM_CIL50}
	\end{subfigure}
	\vspace{-3mm}
	\caption{{\bf Top}: t-SNE~\cite{van2008visualizing} visualization of the features of the Aper algorithm on the CIFAR-100 test dataset with 30 and 50 classes. {\bf Bottom}: t-SNE visualization of the features of \name on the CIFAR-100 test dataset with 30 and 50 classes.}
	\vspace{-3mm}
	\label{fig:benchmark}
\end{figure}

\section{Visualizations}

 In \cref{fig:benchmark}, we present a t-SNE comparison of feature distributions between the Aper and \name on the CIFAR-100 test dataset under different incremental tasks, specifically for 30 and 50 classes. \cref{fig:Aper30} and \cref{fig:Aper50} illustrate the feature distributions of the Aper for 30 and 50 classes, respectively. As new classes are introduced, Aper’s feature distributions become more entangled, with less separation between classes. In contrast, \cref{fig:NCPM_CIL30} and \cref{fig:NCPM_CIL50} show the feature distributions of our proposed \name method under the same settings. Benefiting from guidance by the Neural Collapse geometry, \name achieves better class separation, maintaining a more balanced and distinct feature distribution even as the number of classes increases.
  
  This visualization further confirms the effectiveness of our approach. By aligning the feature space with the Neural Collapse structure, \name dynamically adjusts the feature distribution to maintain inter-class separation and consistency, thereby enhancing adaptability to new tasks and stability to previous ones in the continual learning process. In contrast to traditional methods, this demonstrates that leveraging pre-trained models in combination with NC geometry effectively captures feature evolution in class-incremental learning.

\end{document}